\documentclass{ecai}
\usepackage{times}
\usepackage{graphicx}
\usepackage{latexsym}
\usepackage{verbatim}
\usepackage{algorithm}
\usepackage{algpseudocode}
\usepackage{soul}
\usepackage{todonotes}

\title{Aerial image geolocalization from recognition and matching of roads and intersections}

\author{Dragos Costea \and Marius Leordeanu
\institute{University Politehnica of Bucharest and Autonomous Systems, Romania,
emails: onorabil@gmail.com \and leordeanu@gmail.com} }

\begin{document}

\maketitle

\begin{abstract}
Aerial image analysis at a semantic level is important in many applications
with strong potential impact in industry and consumer use, such as automated mapping,
urban planning, real estate and environment monitoring, or disaster relief.
The problem is enjoying a great interest in computer vision and remote sensing, due to increased computer power and improvement
in automated image understanding
algorithms. In this paper we address the task of automatic geolocalization of aerial images from recognition and matching
of roads and intersections. Our proposed method is a novel contribution in the
literature that could enable many applications of aerial image analysis when GPS data is not available. We offer a complete pipeline for geolocalization, from the detection of roads and intersections, to the identification
of the enclosing geographic region by matching detected intersections to previously learned manually labeled ones, followed by
accurate geometric alignment between the detected roads and the manually labeled maps. We test on a novel dataset with aerial images
of two European cities and use the publicly available OpenStreetMap project for collecting ground truth roads annotations.
We show in extensive experiments that our approach produces
highly accurate localizations in the challenging case when we train on images from one city and
test on the other and the quality of the aerial images is relatively poor.
We also show that the
the alignment between detected roads and pre-stored manual annotations can be effectively used for improving the quality
of the road detection results.
\end{abstract}


\section{Introduction}

The ability to accurately recognize different categories of objects from aerial imagery, such as roads and buildings,
is of great importance in understanding the world from above, with many useful
applications ranging from mapping, urban planning to environment monitoring.
This domain is starting a
flourishing period, as the several technological and computational aspects involved,
both at the hardware and algorithms levels, form in combination very powerful systems
that are suitable for practical, real-world tasks.
In this paper we address two important problems that are not sufficiently studied in the literature.
We are among the first, to our best knowledge, to propose a method for automatic geo-localization
in aerial images without GPS information, by putting in correspondence the real world images
with the publicly available, manually labeled maps from the OpenStreetMap (OSM) project~\footnote{https://www.openstreetmap.org/}.
We solve the task by first learning to detect roads and intersections in aerial images,
and then learn to identify
specific intersections based on a high level descriptor that puts in correspondence
the detected intersections from real world images to intersections detected in the manually labeled OSM maps.
Accurate localization is then obtained by the geometric alignment of the two road maps - the detected ones and the OSM
annotations - at the final step. We present how the alignment to the OSM maps could be used to improve
the quality of the detected roads and intersections.
We also show that
the accurate geometric
registration of roads and intersections can improve both recognition of the roads and the initial localization.
A key insight of our approach is the observation that
intersections tend to have a unique road pattern surrounding them and thus can
play a key role in localization, by
reducing this difficult task
to a sparse feature matching problem followed by a local refined roadmap alignment.
For the accurate detection of roads we use a recent state of the art method~\cite{AlinaNet2016}
that is based on a dual stream local-global deep CNN, which takes advantage of
both the local appearance of an object as well as
the larger contextual region around the object of interest, in order to augment its local appearance and thus improve
recognition performance.

\section{Related work on road detection and localization}

Road detection in aerial imagery has been traditionally addressed by detection methods that use manually designed features
~\cite{kn:mayer2006rereview, lin2012road, laptev2000automatic, klang1998automatic, gruen1995road}.
The recent success of convolutional neural networks~\cite{krizhevsky2012imagenet,simonyan2014very}
has led to greatly improved accuracy and
robust road detection~\cite{kn:mnih2010learning,kn:saito2015building}.
As shown in~\cite{AlinaNet2016}, the lack of good quality aerial images, as well as clutter and occlusion
can greatly affect and significantly degrade the
learning and performance even for top, state-of-the-art architectures.
Post-processing is often required in aerial image analysis~\cite{kn:mayer2006rereview}, but it is not expected
to solve the most difficult cases. There are many approaches proposed for road detection, such as
following road tracks \cite{hu2007road}, local context modeling with CRFs \cite{kn:montoya2014mind}, minimum path methods \cite{turetken2012automated} or using neural networks~\cite{kn:mnih2010learning}.
Arguably, free road vectors are widely available for most of the planet. However,
they are sometimes misaligned and have a poor level of detail.
Therefore some methods attempt to correct these road vectors by aligning them to real rectified aerial images~\cite{Mattyus_2015_ICCV}.
Topological road improvement methods trace back to~\cite{kn:gamba2006improving}. A more recent approach~\cite{kn:montoya2014mind}
uses Conditional Random Fields in conjunction with a minimum cost path algorithm for
improving topology. The authors take into account various cues, such as
context, cars, smoothness between road widths in order to offset road vertices to their real location.
The same authors previously proposed a metric for topology measurement~\cite{kn:wegner2013higher}.

There are several methods related to automatic geolocalization from aerial images, but the tasks they address
differ from ours. Some use known landmarks, others ground floor images or
extra GPS or IMU measurements. Most employ sparse, manually designed features - ours being the first, to the best of our knowledge,
to automatically localize aerial images from recognition and matching of semantic categories, such as roads and intersections,
in the context of deep neural networks. More specifically, related to our work,
geolocalization for unmanned aerial vehicles (UAVs) using sparse manually designed features has been proposed in~\cite{caballero2009unmanned},
while accurate, sub-pixel manhole localization has been proposed using known landmarks~\cite{drewniok1995high}.
A road following strategy for UAVs with lost GPS signal is described in~\cite{frew2004vision}.
Other authors augment a feature-based approach by fusing camera input with GPU and inertial measurement unit (IMU) outputs.
They propose a monocular SLAM approach without visual beacons~\cite{kim2007real,caballero2009vision}, which yields an error of about 5m.
Given the global coverage of aerial images, there has been interest in geolocalizing a ground image using aerial images at training
time~\cite{lin2015learning, workman2015wide, lin2013cross}. Geolocalizing single ground images has also
been recently experimented in~\cite{weyand2016planet}.
An approach loosely related to geolocalization proposed the study of street patterns in order to identify the city class~\cite{kn:barthelemy2008modeling}.

\begin{figure*}
\centering
\includegraphics[height=10cm]{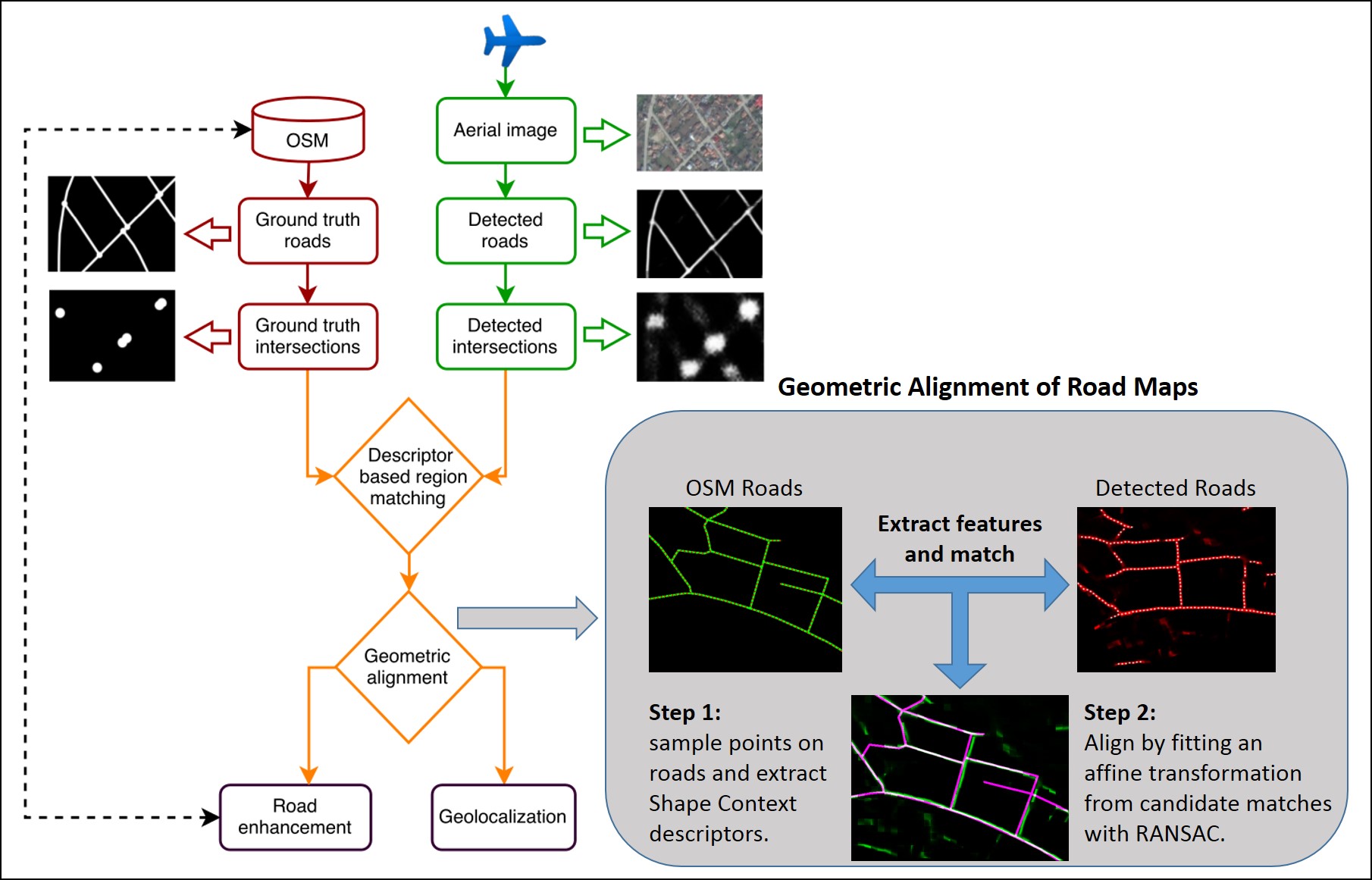}
\caption{Framework overview: ground truth road maps and intersections are extracted from OSM, with known locations (left pathway). High-level descriptors are
extracted from each intersection and stored offline.
At test time,
roads and intersections are automatically detected in aerial images using the dual stream CNN model~\cite{AlinaNet2016}.
The same type of intersections descriptors are extracted and matched against the OSM set of descriptors in order to localize a given detected
intersection in the aerial image. This provides an initial localization that
is further improved by geometric alignment. This also helps in intersection identification -
only pairs of intersections with high alignment score are put in correspondence.
}
\label{fig:overview}
\end{figure*}

\section{Our approach}

Our method has several stages: 1) road pixelwise classification in a given aerial image; 2)
detection of intersections based on the detected roads; 3)
identification of a given intersection by matching its surrounding region to regions from a
stored dataset of OpenStreetMap(OSM) road and interesections maps. At this stage we
keep, for each test intersection, a list of closest OSM interesections in the intersections descriptor space;
4) accurate geometric alignment for improved localization and road detection enhancement.
At this stage we keep from the list of candidate intersection matches the one with minimum
geometric alignment error.
In this work we focus on recognition and localization of
given detected intersection. We use intersections as anchors for localization
for three reasons. First, once intersections are found and images are aligned to known roadmaps
the location of any given point in the image follows immediately. Second,
intersections are sparse and require very little computational and storage costs
for recognition and matching. Third,
they are also sufficiently discriminative localization when their surrounding area is taken into account.
They tend to have a unique pattern of roads
in the neighborhood region,
which acts as a unique fingerprint that is useful for location recognition.
We present an overview of our approach in Figure~\ref{fig:overview}.
Note that while we did not use any GPS information for localization, we assumed that we know the orientation of the image
with respect to the cardinal points - an information that is easily obtained
with a compass in a real world situation.
To account for small errors in orientation estimation we added a random Gaussian noise to the test
image rotation angle with 0 mean and standard
deviation of 5 degrees. While the added noise
affected slightly the performance of intersection recognition, it did not influence the final geometric alignment stage
that is affine invariant.
We detail the stages of our pipeline next.

\subsection{Finding roads and intersections}

\begin{figure*}
\centering
\includegraphics[height=8cm]{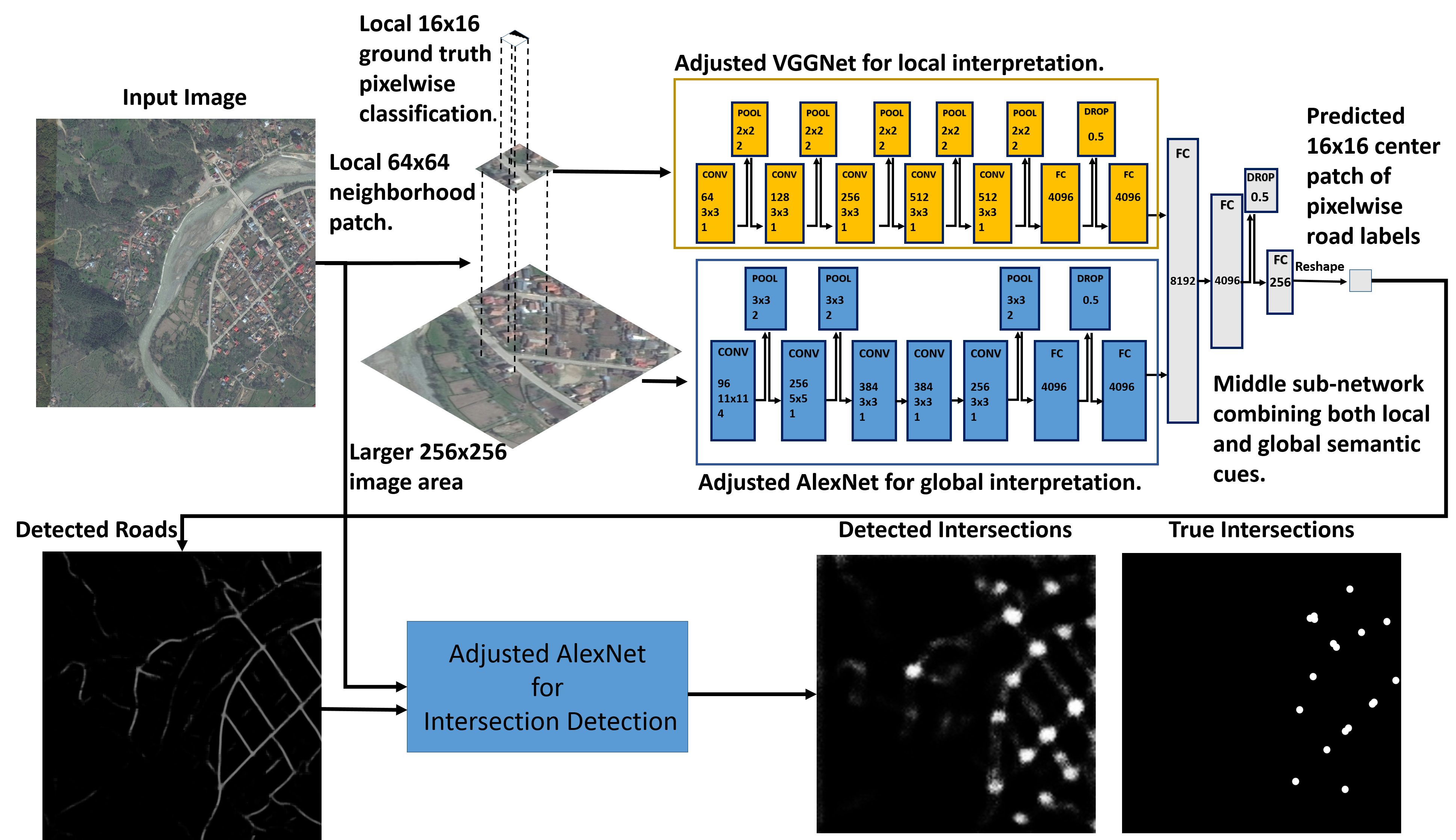}
\caption{Our system for detection of roads and intersections. We first detect roads in the image
by classifying each individual pixel using the recently proposed LG-Net model, that processes information
along two pathways - a local one for reasoning based on local appearance, and a global pathway for image interpretation
at the level of the larger contextual region. The detected roadmap is then passed to an adjusted AlexNet model trained
for the task of intersection recognition. Intersections are detected by a scanning window approach followed by non-maxima suppression.
}
\label{fig:roads_and_intersection_detection}
\end{figure*}

\paragraph{Detection of roads:}
We train a state-of-the-art dual stream local-global Convolutional Neural Network~\cite{AlinaNet2016} (LG-Net)
on the task of road detection (Figure~\ref{fig:roads_and_intersection_detection}).
The network combines two pathways, one based on an adjusted VGG-Net~\cite{simonyan2014very}
that uses local appearance information (a local 64x64 patch surrounding the road region)
and the other, based on an adjusted AlexNet~\cite{krizhevsky2012imagenet}, which takes as input a significantly larger
neighborhood (256x256) for contextual reasoning. The two pathways are joined in the last FC layers and the output is a small 16x16 center
patch having 1's for road pixels and zeros otherwise. The final road map is obtained by dividing the larger aerial images into disjoint 16x16 patches, which are
classified independently. In the experiments presented in~\cite{AlinaNet2016}
the local-global network achieves an F-measure that is consistently superior
to a network that has only the local pathway. Also, compared to previous contextual approaches to road detection,
ours avoids hand crafted cues,
such as the nearby cars and consistent road width~\cite{Mattyus_2015_ICCV} or nearby lines~\cite{yuan2013road},
and effectively learns to reason about context by considering the larger area containing the road.

\paragraph{Detection of intersections:}
For the detection of intersections we trained an adjusted AlexNet architecture, modified to output a single class to signal
the presence or absence of an intersection at a given point in the image. We considered as input several channels containing
the original RGB image as well as the estimated roadmap provided by the LG-Net.
Including the channels with the original RGB low level signal improved the maximum
detection F-measure from $65.18\%$ to $67.71\%$, in our experiments, using a scanning window approach with non-maxima
suppression.
The most relevant of the two
types of input is the estimated roadmap that represents signal at a higher, semantic
level of image interpretation. Note
that intersections, by definition, are directly related to the existence of at least two roads that intersect.
In order to speed up the detection of intersections we classified pixels on the grid (with steps of 10 pixels)
and obtained the final dense intersections map by interpolation. This resulted in a speedup
by two orders of magnitude at
the cost of a relatively small
decrease in detection quality. In Figure~\ref{fig:roads_and_intersection_detection},
we also present the system for intersection detection with an example estimated map
of intersections. We notice that most intersections are detected, while, in some cases, intersections
seem to be correctly detected in the image but are not present in the OSM, which we considered as ground truth.
Note that such inconsistencies between images and manually labeled roads
are not uncommon in OSM.

\subsection{Automatic geolocalization}

We represent each intersection by a descriptor which is learned such that identical intersections from detected roads and OSM roads
should have similar descriptors,
while descriptors for different intersections should be as far separated as possible.
For extracting the intersection descriptors we start from the
modified AlexNet trained for intersection detection, such that
the last FC layer of 4096 elements
is used as a descriptor. Intersections from the detected road maps will be matched against
a database from OSM using Euclidean distances in descriptor space. While this approach proves to be very
effective, we further improve the performance by
fine-tuning the network using backpropagation for adjusting distances in descriptor space in order to improve matching performance.
(Figure \ref{fig:performance}).
Localization is further refined by the geometric alignment between the estimated roads and the OSM roads in the regions centered at the intersections
that have been put in correspondence. We detail next the algorithms for matching and localization.

\paragraph{Descriptor extraction and learning:}
We extract descriptors for intersection images in a way that is similar to~\cite{lin2015deep}.
Moreover, we fine-tune the descriptor extracted for intersections from the neural network, so as to
minimize the distance between identical intersections and maximize the distances between dissimilar ones.
First, we train the modified Alexnet for intersection detection. Second
we fine tune the network weights in a Siamese-like fashion,
with corresponding intersection pairs from estimated roadmaps and OSM, respectively,
marked as positive and different intersection pairs marked as negative.
See~\cite{hadsell2006dimensionality} for details on this type of training.
The robust loss formula we use takes in consideration the ground truth label $y$,
which is $1$ if the intersections are the same and $0$ otherwise,
the squared Euclidean distance $d$ between pairs of intersections descriptors and a margin $m$,
which gives zero penalty to descriptors $\mathbf{a}$
and $\mathbf{b}$
from different intersections that are at a distance of at least $m$
in descriptor space:

\begin{equation}
L(y) = \frac{1}{2}y d + \frac{1}{2}(1-y) \max(m-d, 0)).
\end{equation}

\paragraph{Intersection identification:}
The learning phase
creates a descriptor for each intersection image. Similar images will correspond to descriptors that are close in Euclidean space.
When matching two regions centered at two candidate intersection matches, we also consider the descriptors of the nearby intersections.
This results
in a bipartite graph matching problem for matching two sets of descriptors.
It is possible, as nearby intersections usually have similar regions to wrongly match detected
intersections to their neighbor OSM intersections
,  but such local misplacements
are most often fixed at the final geometric alignment step when all the roads details in a region
are taken into account. Next
we present our method for finding correspondences between detected
intersections and the ones from OSM, by matching sets of intersections from their corresponding regions. These
neighborhoods of a certain radius centered at the intersections of interest. As our experiments show, the larger
this radius the more accurate the intersection identification. This is expected, as larger
regions include more road structures that are unique to a specific urban area.

\begin{algorithm}
\caption{Intersection identification by matching regions}
\label{alg:bijscore}
\begin{algorithmic}
\For{\textbf{each} road detected test intersection $i_T$ and given radius $w$ }
\State Gather roads and intersections from region intersection region $R_w(i_T)$.
\For{\textbf{each} label OSM intersection $i_L$}
\State Gather roads and intersections from region $R_w(i_L)$.
\State \textbf{Compute matching distance between regions}:
\State $\;\;\;$ 1) Get nearest neighbor distance $t_{j}$ between \\
       $\;\;\;\;\;\;\;\;\;\;\;\;\; \; \; \;$ each intersection $i_j$ in $R_w (i_T)$ to intersections from $R_w (i_L)$.
\State $\;\;\;$ 2) Compute sum of 1NN distances $S_t(i_T, i_L)  = \sum_{j}t_{j}$.
\State $\;\;\;$ 3) Get nearest neighbor distance $l_{j}$ between \\
       $\;\;\;\;\;\;\;\;\;\;\;\;\; \; \; \;$ each intersection $j$ in $R_w (i_L)$ to intersections from $R_w (i_T)$.
\State $\;\;\;$ 4) Compute sum of 1NN reverse distances $S_l(i_L, i_T) = \sum_{j}l_{j}$.
\State $\;\;\;$ 5) Set distance between intersections: $d(i_T, i_L) = (S_t(i_T, i_L)+S_l(i_L, i_T))/2$.
\EndFor \\
\Return list $L_k(i_T)$ of $k$ closest $i_L$'s OSM intersections to $i_T$ using distance $d(i_T, i_L)$.
\EndFor
\end{algorithmic}
\end{algorithm}

\paragraph{Geometric alignment:}
Although a location can be theoretically determined by a single correctly identified
intersection and a correct rotation with respect to the cardinal points
, in order to have a robust match and further improve the initial
localization (which could be off due to intersection detection misalignments),
we also estimate for a given pair of candidate intersection matches $(i_T,i_L$,
a geometric affine transformation
between the roads in regions $R_w(i_T)$ and $R_w(i_L)$
Then, a misalignment measure is computed such that
most outlier candidates in the list $L_k(i_T)$ of a given test intersection (found using Algorithm 1)
$i_T$ are removed.
The 2D registration procedure is performed by
sampling road points from the test and query images and computing Shape Context descriptors\cite{belongie2000shape}
at sampled locations. Using kNN with Shape Context descriptors, a list
of candidate correspondences are found
and an affine transform is robustly estimated using RANSAC.
Then, the Euclidean distance transform ($bwdist$ Matlab function)
is used in order to compute the symmetrized Chamfer distance
between the two registered roadmaps, as a measure of misalignment - which, in practice
yields significantly better results.
Other approaches (such as \cite{Mattyus_2015_ICCV}) also
proposed road alignment. Ours is fast and very effective for
rejection of outlier intersection matches,
improving localization and road enhancement (next Section).
The more detailed overview of our localization algorithm is presented below:

\begin{algorithm}
\caption{Geolocalization algorithm}
\label{alg:geoloc}
\begin{algorithmic}
\For{\textbf{each} road intersection $i_T$}
\State 1) Find list $L_k(i_T)$ of $k$ candidate matches $i_L$ from OSM using Algorithm 1.
\State 2) Compute $k$ symmetric Chamfer distances $C(i_T, i_L)$ between \\
$\;\;\;\;\;\;\;\;\;\;\;\;\; \; \; \;$ region $R_w(i_T)$ and the corresponding regions $R_w(i_L)$ of each $i_L \in L_k(i_T)$.
\State 3) Return aligned $i_L^{*}$ from $L_k(i_T)$ with minimum distance $C(i_T, i_L^{*})$.
\EndFor
\end{algorithmic}
\end{algorithm}

\subsection{Enhancing the road map}

We can use the aligned OSM roadmaps to improve the detected roads and vice-versa - since OSM roadmaps
sometimes contain wrongly labeled roads, or do not reflect recent road changes. Here present
a simple but effective method: 1) we apply a soft dilation procedure on the estimated roadmap
and multiply it, pixel by pixel, with the aligned OSM map; 2) the resulted soft output is then smoothed
with a Gaussian filter and the result is thinned using a standard
nonmax suppression method for boundary detection. 3) after thinning the roads are dilated back, to achieve
the initial thickness. The results are substantially better, as expected, greatly improving the
similarity between the roads found and the OSM roads - the f-measure in road detection improved from
$66.5\%$ to $93.9\$$. \textbf{Important note:} this procedure does not use ground truth localization, but only the
entire OSM dataset and relies on the accuracy of the automatic matching and alignment algorithms. It has proved
generally effective even
when the localization was wrong but the road structure between the matched OSM region and the test image
was similar. We present qualitative results
in Figure \ref{fig:enhanced_roads}.

\section{Experimental analysis}

\begin{figure*}
\centering
\includegraphics[height=8.5cm]{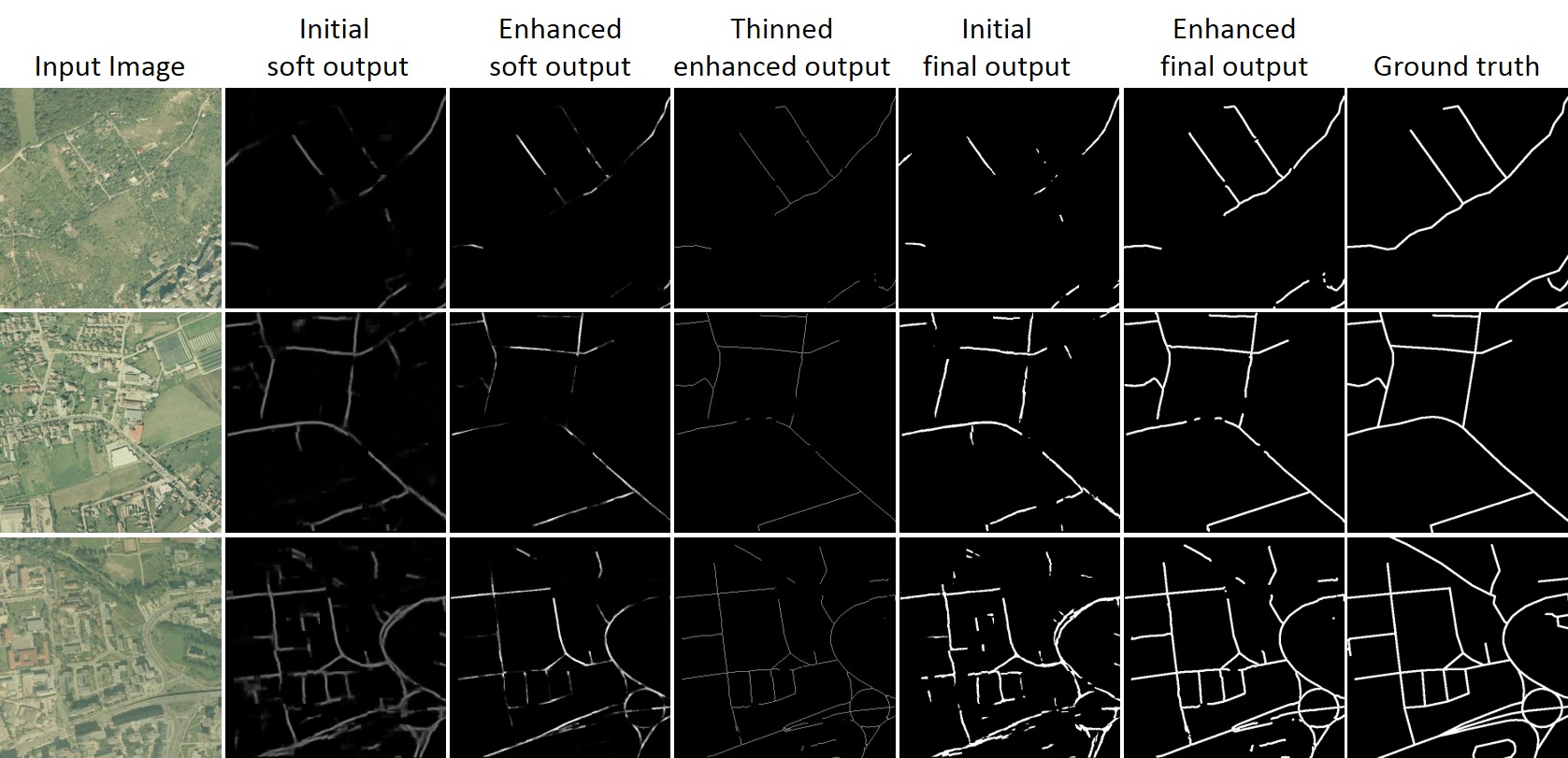}
\caption{Enhancing the road detection by region recognition and
geometric alignment to OSM roads. Our simple procedure, described in the text,
could be useful for both improving the detected roadmap in the test image and correcting
the OSM manually labeled maps. Note that for road enhancement we used the automatically matched
and aligned regions from OSM using the initial estimated roadmaps,
and NOT the ground truth matches.
}
\label{fig:enhanced_roads}
\end{figure*}

\paragraph{Two Cities Dataset:}
We collected aerial images of two European cities (termed A and B)
and automatically aligned them with the OSM road maps for training and evaluation.
We plan to make the dataset public.
The images are 600x600px, have the spatial resolution of 1m/pixel and cover an area of about 70 sq. Km
each. We use city A for training and validation and images from city B for
testing. The quality of the images is
fairly low, which makes the task of road detection and localization very challenging, even for the human eye (see example
images in Figure~\ref{fig:enhanced_roads}).

\begin{figure}
\centering
\includegraphics[height=8.1cm]{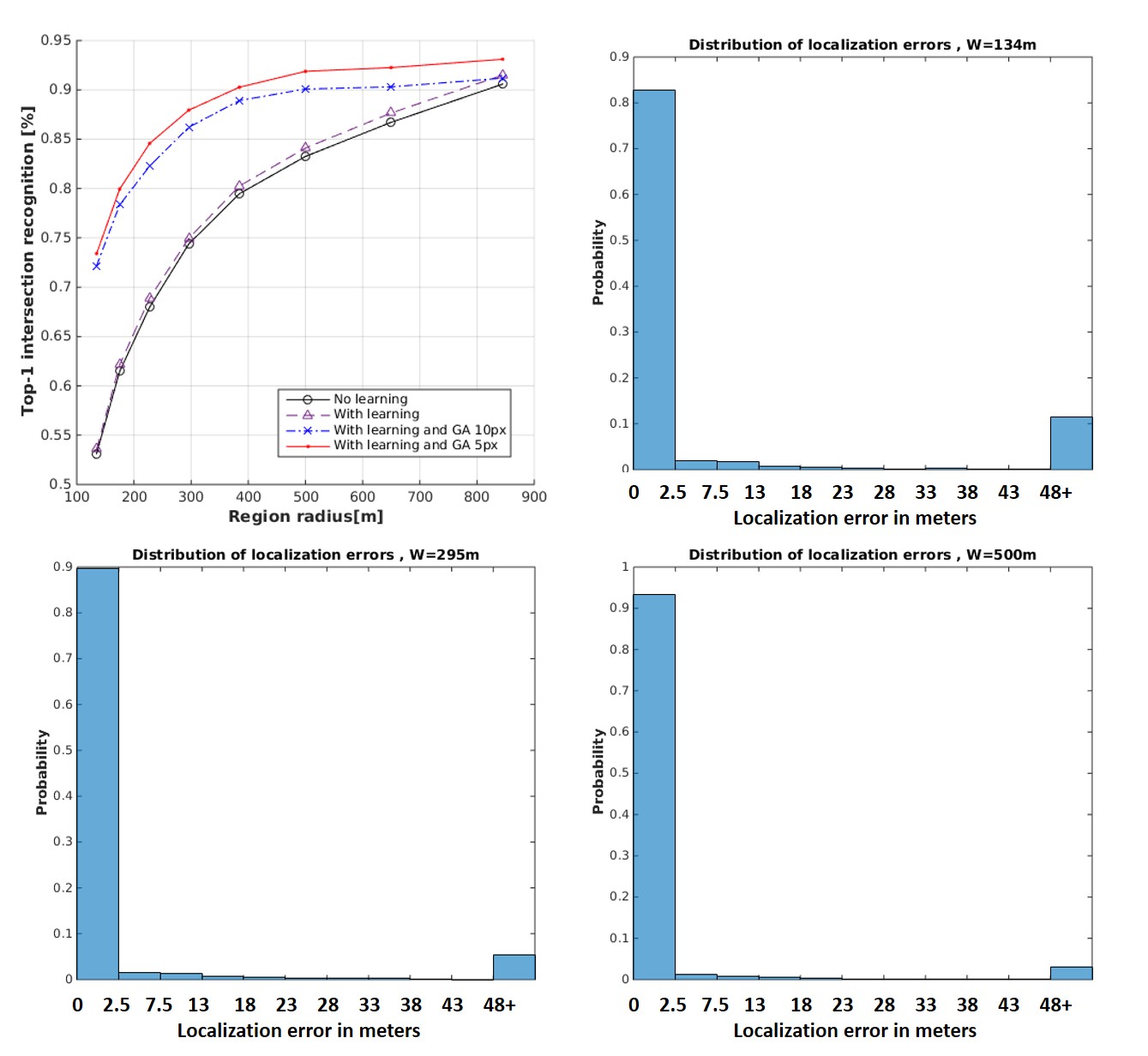}
\caption{Performance evaluation. Top left plot: performance increases with the region radius. Note that intersections descriptor learning as well as the final geometric alignment method significantly improve localization accuracy. The other three plots, showing distribution of errors per distance in meters show that our approach is able to correctly localize an intersection with an error of maximum 2.5 meters in at least $90\%$ of cases.}
\label{fig:performance}
\end{figure}

Figure~\ref{fig:performance}
presents the average performance measures after
geolocalizing all 3177 intersections from city B.
We present intersection identification (recognition) rates versus the region radius (top left plot). As expected performance increases
as the region radius increases, at the cost of more computation and data being required. We also demonstrate
that the geometric alignment phase significantly increase performance, bringing it close to the $90\%$ mark even when the
region radius is small. The plot also presents the consistent improvement brought by fine tuning the descriptors to optimize
intersection matching. The other three plots present the distribution of localization
errors in meters. We notice that most errors (around or above $90\%$ of them)
are below 2.5 meters, that is below 3 pixels for the image
resolution available in our experiments. This error is very small considering the poor image quality and
the errors present in the OSM itself, which was considered as ground truth.
For these reasons we believe that our results demonstrate
high level of localization accuracy for our system,
which could be very effective
in most cases when the GPS signal is lost.

\paragraph{Computational details:}
Training time for road detection and intersections descriptor learning took between 3-5 days on a GeForce GTX 970
GPU with 4Gb memory and 1664 CUDA cores. At test time,
road extraction speed is 5km\textsuperscript{2}/s, at a spatial resolution of 1m/pixel and represents
the most expensive task for geolocalization. Intersection detection takes 0.7km\textsuperscript{2}/s, while localization by means of kNN
in intersection descriptor space and
geometric alignment is an order of magnitude faster in the context of searching within the limits of a $70$ sq. Km
city.

\section{Discussion and Conclusions}

We have presented a complete system for geo-localization from aerial images
in the absence of GPS information. Our proposed pipeline includes many contributions with efficient methods for
road and intersection detection, intersection recognition with geometric alignment for accurate localization, followed
by road detection enhancement. There are many potential applications for our approach in areas such
as urban planning, tracking structural changes, updating of existing maps and environment monitoring.
Our system could also be used in the context of
unmanned aerial vehicles, in order to correct their GPS localization or to make their flight possible even
when GPS signal is lost.
We estimate that if the search area was only $5$ times smaller than in our experiments,
the automatic localization
would be tractable for onboard processing, in near real-time, for current generation of NVIDIA's embedded GPUs (Jetson TX1).
For nighttime use for example,
the roads are generally 'extracted' by means of street lightning, which makes the
problem of road and intersection detection easier - thus even
more accessible for on-board processing.
We have proven that geolocalization from images alone, using learned
high level features is feasible and can achieve a high level of accuracy. It
can be used as a GPS alternative or in conjunction with GPS, bringing valuable contributions
to the literature and also to
many applications that require
offline or online, realtime processing.

\ack The authors would like to thank Alina Marcu for his dedicated assistance with
some of our experiments. Marius Leordeanu was supported in part by
CNCS-UEFISCDI, under project PNII PCE-2012-4-0581.

\bibliography{complete} 
\bibliographystyle{ecai}
\end{document}